# On the Importance of Diversity in Re-Sampling for Imbalanced Data and Rare Events in Mortality Risk Models

Yuxuan, Y, Yang

School of Computing and Information Systems, The University of Melbourne, yyxdiana.yang@gmail.com

Hadi Akbarzadeh, HA, Khorshidi

School of Computing and Information Systems, The University of Melbourne, hadi.khorshidi@unimelb.edu.au

Uwe, U, Aickelin

School of Computing and Information Systems, The University of Melbourne, uwe.aickelin@unimelb.edu.au

Aditi, A, Nevgi

Metabolic Medicine, The University of Melbourne, aditi.nevgi@hotmail.com

Elif, E, Ekinci

Metabolic Medicine, The University of Melbourne, elif.ekinci@unimelb.edu.au

Surgical risk increases significantly when patients present with comorbid conditions. This has resulted in the creation of numerous risk stratification tools with the objective of formulating associated surgical risk to assist both surgeons and patients in decision-making. The Surgical Outcome Risk Tool (SORT) is one of the tools developed to predict mortality risk throughout the entire perioperative period for major elective in-patient surgeries in the UK. In this study, we enhance the original SORT prediction model (UK SORT) by addressing the class imbalance within the dataset. Our proposed method investigates the application of diversity-based selection on top of common re-sampling techniques to enhance the classifier's capability in detecting minority ('mortality') events. Diversity amongst training datasets is an essential factor in ensuring re-sampled data keeps an accurate depiction of the minority/majority class region, thereby solving the generalization problem of mainstream sampling approaches. We incorporate the use of the Solow-Polasky measure as a drop-in functionality to evaluate diversity, with the addition of greedy algorithms to identify and discard subsets that share the most similarity. Additionally, through empirical experiments, we prove that the performance of the classifier trained over diversity-based dataset outperforms the original classifier over ten external datasets. Our diversity-based re-sampling method elevates the performance of the UK SORT algorithm by 1.4%.

## 1 Introduction

There were over 2.3 million surgical procedures between 2017-2018 in Australia [1]. This statistic is associated with the realization that in an ageing population, a significant proportion of these patients have comorbid conditions, resulting in more complex surgeries, escalating surgical risk and heightened risk of post-surgical complications and treatment [2]. All of the above has led to the creation of numerous risk stratification tools, aimed at classifying and predicting post-operative mortality rates for various major elective surgeries. Some examples include the Physiological and Operative Severity Score for the enumeration of Mortality and morbidity (POSSUM) [3] and the Surgical Outcome Risk Tool (UK SORT) [4], which was subsequently developed in 2014 in the UK and has shown superior performance. SORT calculates a percentage mortality risk for patients undergoing surgery.

In this study, we enhance the Precision-Recall Area under the curve (PR-AUC) of the UK SORT model by utilizing data collected from the Austin Health Research Ethics Committee (Austin dataset) as approved by the Austin Health

Human Ethics Committee. The Austin dataset used in this study contains admissions of surgical inpatients into Austin Health, a tertiary teaching hospital in Melbourne, Victoria, from mid-2013 to mid-2016. The sample data contains 1853 records after cleaning. Each record contains information on age, sex, type of surgery, elective or emergency status of surgery, Charleston Comorbidity Index (CCI), and comorbidities. To the best of our ability, we have attempted to mirror each of the input variables in the original UK study to their corresponding Australia-equivalent metrics. It is important to point out that the ASA-PS score (used in UK SORT) is not explicitly documented in the Austin dataset, and therefore, we have translated CCI scores to ASA-PS scores to the best of our ability. More specifically, CCI 0 is translated to ASA-PA1; CCI 1 is translated to ASAPS 2; CCI 2 is translated to ASAPS 3; CCI 3 is translated to ASA-PS 4; and CCI 4 and above is translated to ASAPS 5. The UK SORT model has three categories of urgency of the operation (expedited, urgent, immediate), while the Austin dataset has two categories called elective and emergency.

Given mortality is an extremely rare event for most surgeries, we noticed that the data collected is significantly biased towards surviving patients with a ratio of 1974:59 (surviving vs. deceased). We refer to such scenarios as between-class imbalanced distributions of data. It is a common occurrence in medical studies such as disease detection studies. Imbalanced classes usually occur due to the sparsity of a particular class of labels. This can be described simply as a scenario whereby the number of instances labelled as "Positive" (minority class) is significantly lower than those labelled as "Negative" (majority class).

Extremely imbalanced datasets typically denote a scenario where the imbalance ratio of a dataset is extremely skewed towards the minority class. The sample data we received from Austin is extremely imbalanced with a 3% mortality rate (minority class). The impact of such underlying characteristics is that any subsequent learning classifier may encounter issues when the minority instances are regarded as anomalies or noise within the dataset [5]. Imbalanced datasets therefore have huge implications for classifier performance and studies have suggested this occurrence poses a significant challenge to most learning algorithms, as most of these algorithms emphasise on prediction accuracy rather than ability to predict minority classes [6]. This means that a lot of widely used classifiers have produced unsatisfactory classification results in terms of predicting the minority class. As a result, there has been rising interests in existing research to improve prediction accuracy on minority classes in the event of imbalance class distribution [7].

The aim of this study is to enhance the capability of the UK SORT model in predicting minority classes by re-training it over a synthetised dataset based on our proposed diversity-based selection approach. Our proposed model produces a more indicative and accurate result of the associated mortality risk based on a myriad of patients' pre-existing conditions. This study particularly contributes to better decision-making when handling imbalanced datasets, such as in understanding the mortality risk of major surgeries amongst surgeons and prospective patients and delivering higher level of treatment outcomes.

We propose a diversity-based algorithm that can be used as a drop-in function for a variety of re-sampling techniques. The first phase of our experimental design is to incorporate the proposed diversity-based algorithm to train and evaluate the performance of various classifiers on a selection of medical and non-medical datasets obtained from the UCI repository [8]. We also use "Oil", a benchmark dataset used to detect oil spills from satellite images, as obtained from [9] [10]. We subsequently apply the proposed diversity-based algorithm, coupled with three re-sampling techniques and three re-sampling ratios to explore whether the same approach will improve the predictive performance of UK SORT.

The rest of paper is organised as follows: Section 2 describes related work and a literature review. Section 3 explains the methodology of our diversity-based sampling drop-in functionality. Section 4 provides an overview of our experimental design and its performance when validated against a list of imbalanced datasets. Section 5 provides an explanation of our case study results. In Section 6, we conclude our findings, highlight limitations of our study and discuss future research.

## 2 Related Work

The current literature addressing imbalanced dataset concentrates on three main areas, i.e. "cost-sensitive learning", "algorithmic/ensemble method" and "re-sampling techniques".

Cost-sensitive learning is a typical technique which can be utilised to finetune the learning classifier. The crux of its implementation is the underlying assumption in attributing a higher cost to the mislabelling minority class data. It is a versatile approach which enables users to allocate different costs scaled to the minority class sizes within the dataset or at the algorithm level. Ensemble-based methods are usually applied to improve the outcome of minority classification by combining multiple base classifiers together to form a single classifier which collectively outperforms each of the independent base classifiers included. Both cost-sensitive learning and ensemble methods are popular areas. For instance in [11] a cost-sensitive component is embedded in the objective function of Neural Network (NN) learners and achieves a statistically significant improvement of results.

A typical example of an algorithmic/ensemble method, is the approach described in [12]. Multi-objective optimisation methods are employed to guide an ensemble of other machine learning methods to impute missing data in an efficient manner. Results show that performing imputation simultaneously with classification does indeed improve performance.

Re-sampling is the main focus of our study. Re-sampling techniques are often employed when a dataset is imbalanced with the objective to create a more balanced dataset. Re-sampling can broadly be undertaken in one of three main approaches, namely over-sampling, under-sampling and a combined approach (hybrid methods).

There are two main categories of over-sampling techniques: Random over-sampling and synthetic over-sampling such as the Synthetic Minority Over-Sampling Technique (SMOTE) [13].

Random over-sampling involves a random selection with replacement of minority class items within a dataset. Once an instance has been randomly selected, it is duplicated and added into the training dataset. This approach is repeated until the desired ratio of majority and minority class is achieved. This method is computationally efficient and can be readily implemented as it does not depend on any inherent knowledge or understanding of the underlying data [7].

SMOTE is a popular synthetic over-sampling technique where the minority class data are not duplicated, but rather extra synthetic samples of the minority class are generated [13]. The creation of synthetic minority class data takes place at the "feature level" and not at the "data level". The synthetic minority class items are created by randomly choosing specified k-nearest neighbours of a minority class sample and applying a rand(0,1) distribution to the distance between the two points. This results in a randomised point which lies between the two examples. The result of the original study suggests that SMOTE improves performance due to the expansion of decision regions which contains nearby minority data points, rather than simply duplicating minority samples which narrows the decision region with contrasting effects [13].

Subsequent research has proposed various extensions, such as Borderline-SMOTE [14], DBSMOTE [15] or Geometric SMOTE (G-SMOTE) [16]. Most recently, G-SMOTE is proposed as an enhancement to the original SMOTE synthetic data creation process. The approach taken is to utilise a flexible geometric region created by an algorithm with parameterised inputs. Synthetic data is then generated within the confines of the specified geometric region. The main objective of this approach is to prevent the generation of synthetic minority data which are noisy and to increase the variety of minority class data within the expanded geometric region. It is regarded as a "drop-in" function for the traditional SMOTE method and results of the study suggest that G-SMOTE is suitable for datasets with higher imbalanced ratios. Further research [16] has attributed the success of G-SMOTE to the capability of the algorithm in generating synthetic data points and improving diversity of generated instances within the safe region of the input space.

More recent studies have identified that traditional over-sampling methods (e.g. SMOTE) may be limited by their tendency to casually generate synthetic instances which encroaches into the input region of the majority class instance. This thereby impairs the performance of the subsequent learning classifier [17; 18]. These studies have subsequently focused on improving the diversity of the minority class data while simultaneously maintaining the integrity of the minority and majority sample regions.

Sharma *et al.* [18] propose a probability density-based approach for majority class instances and their study concludes by stating that the "SWIM" approach is superior when evaluated with extremely imbalanced data.

Bennin *et al.* [17] propose an alternative sampling approach which utilises the concept of Mahalanobis Distance (MD), in conjunction with an inheritance heuristic and genetic algorithms. The idea is to generate newly synthesised minority instances that are unique by using two relatively distant parent instances. Thus creating children data that are different to their parents (i.e. the existing minority class).

The study by Bennin *et al.* [17] also highlights the importance of diversity-based selection by stressing the difficulty in common over-sampling approaches (e.g. random over-sampling, SMOTE, Borderline-SMOTE) in first determining the appropriate *k* parameter and subsequently increased likelihood of the creation of near-identical or completely duplicated instances. This is in addition to an earlier study which was conducted in [19], where it is illustrated that the k-nearest neighbours approach over-generalise the sampling approach by considering all the k-nearest neighbours and not the minority class specifically. The effects of lack of diversity in the results of classifier can also result in an increase in higher false positives (minority labels), which may be costly in commercial settings as it requires significant resources to evaluate wrongly classified predictions.

In this paper, we propose to build on the "drop-in" functionality from G-SMOTE and offer a new diversity-based strategy which can be utilised to artificially bolster diversity of the minority instances while allowing users to preserve and apply any of the existing mainstream re-sampling techniques highlighted above. The proposed method is extremely flexible and can be easily implemented.

# 3 Methodology

In this section, we describe our proposed drop-in function that maintains diversity in re-sampling. Our function selects the subset of instances that have the highest diversity using the Solow-Polasky measure [20]. To calculate the Solow-Polasky measure, first we need to construct a distance matrix ($M = \begin{bmatrix} m_{ij} \end{bmatrix}$). We use Euclidean distance to measure the distance between instances. Then, we obtain the inverse of the distance matrix ($M^{-1}$). Finally, the diversity measure (D(S)) is the summation of the elements of $M^{-1}$ as formulated in equation 1.

$$x = \frac{-b \pm \sqrt{b^2 - 4ac}}{2a} \quad (1)$$

where $d(s_i, s_j)$ denotes the distance between instances of dataset S.

The Solow-Polasky measure has three properties that guarantee the effectiveness of this diversity measure [21]. These properties are (a) Monotonicity in varieties: the diversity measure increases by adding an individual instance that was not in the dataset, (b) Twining: the diversity measure remains the same by adding an individual instance that was already in the dataset, (c) Monotonicity in distance: the diversity of dataset S should not be smaller than another set data $S'$ if all pairs in S are as distant as all pairs in $S'$.

To select the most diverse subset of instances, we should examine all possible subsets. However, examining all possible subsets is computationally expensive. Thus, we use a greedy algorithm to find and discard instances with the least contribution to the diversity of the dataset. The contribution of an instance is defined as how much the diversity of the dataset reduces if that instance is removed. To calculate the contribution of each instance, we need to obtain the inverse of the distance matrix with and without that instance. The distance matrix without the instance can be shown as A, where M is $\begin{bmatrix} A & b \\ b^T & c \end{bmatrix}$. So, $M^{-1}$ can be formulated as $\begin{bmatrix} \overline{A} & \overline{b} \\ \overline{b}^T & \overline{c} \end{bmatrix}$ where c and $\overline{c}$ are single elements, b and $\overline{b}$ are vectors and $b^T$ and $\overline{b}^T$ are their transpose. the contribution of an instance can be formulated using equation 2 as proven in [22]. Using this equation, there is no need to construct the inverse matrix two times and consequently we reduce the computation cost. Algorithm 1 shows how the diversity-based selection works using a greedy algorithm.

$$\sum M^{-1} - \sum A^{-1} = \frac{1}{\overline{c}} \left( \sum \overline{b} + \overline{c} \right) \quad (2)$$

---

ALGORITHM 1: Diversity-based selection

**Input:** A set of instances *S*; The number of instances we want to keep *r*
**Output:** A set of diverse instances $S_{Div}$

$N \leftarrow$ The number of instances in $S$
$S_{Div} = S; d = 0; C = $ Null
**While** $d < N-r$ **do**
  $N_{Div} \leftarrow$ The number of instances in $S_{Div}$
  **for** $i$ 1 **to** $N_{Div}$ **do**
    $C_i \leftarrow$ Calculate the contribution of instance $i$ to the diversity using equation 2
  **end for**
  $n \leftarrow \underset{i}{\operatorname{argmin}} C$ /*Find the instance with minimum contribution*/
  Remove $n$th instance from $S_{Div}$
  $d = d + 1$
**end while**

# 4 Validation

## 4.1 Experimental Design

We validate the proposed diversity-based algorithm against an assortment of ten medical and non-medical imbalanced datasets, with varying dimensions and imbalance ratios. The datasets and their characteristics are described in Table 1, and "Ratio" is used to indicate the proportion of majority to minority instances within each dataset.

Each specific dataset has been randomly categorised into training and testing datasets using a 75:25 split respectively. The performance of our proposed method is then evaluated against a benchmark re-sampling method without diversity-selection, namely under-sampling, over-sampling and SMOTE. For the computation of the SMOTE benchmark, we have retained the default value k=5 parameter within an R package [23] as it has been proven that a value of 5 is the optimal value [18] for practical applications.

**Table 1: Dataset characteristics**

| Dataset | Name | Dimension | Size | Ratio | Dataset | Name | Dimension | Size | Ratio |
|---|---|---|---|---|---|---|---|---|---|
| D1 | Diabetes | 8 | 768 | 65-35 | D6 | Oil | 49 | 937 | 95-05 |
| D2 | Vowel 0 | 10 | 990 | 91-09 | D7 | Vehicle Bus | 18 | 846 | 74-26 |
| D3 | Vowel 2 | 10 | 990 | 91-09 | D8 | Pima | 8 | 768 | 65-35 |
| D4 | Wisconsin | 9 | 683 | 65-35 | D9 | Wine White Low vs High | 11 | 1,243 | 85-15 |
| D5 | KDD Control | 60 | 600 | 83-17 | D10 | Wine Red Low vs High | 11 | 280 | 77-23 |

We develop classifiers on training data and measure the performance on the test data. Each procedure is repeated 30 times for ten datasets over three imbalance levels using four classifiers in a bid to mitigate the impact of randomness in the dataset splitting and sampling processes. The classifiers are Generalised Linear Models (GLM), Decision Trees (DT), k-Nearest Neighbour (KNN), and Random Forest (RF). We chose KNN and RF as they are sensitive to imbalanced data based on their model assumptions [24]. GLM is a commonly used classifier in medical problems, even if the data is imbalanced [4]. It is also an effective classifier when classes are linearly separable. DT work based on developing decision regions which are influenced by re-sampling methods [25].

We evaluate the performance of the classifiers on test data using PR-AUC (in order to address issues with the inherent bias of global performance metrics like ROC-AUC). Studies have suggested that ROC-AUC may be too optimistic when there are low false positive rate, which is typical when classes are imbalanced [26]. It is also shown in [27] that ROC-AUC suffers from sensitivity issues to class distributions and error costs. Separate research has also shown that PR-AUC is more informative than ROC-AUC in evaluating binary classifiers performance as it emphasises the accuracy of future positive predictions by evaluating the fraction of true positives among all positive predictions [28].

In PR-AUC, Recall is the proportion of correctly predicted positive instances to all instances in the positive class (3). Precision is the proportion of correctly predicted positive instances to all predicted positive instances (4). In the following equations (3) and (4), TP denotes "True Positive" (i.e. the number of instances from positive classes predicted correctly), FN denotes "False Negative" (i.e. the number of instances from positive classes predicted negative) and FP denotes "False Positive" (i.e. the number of instances from negative classes predicted positive). Furthermore, PR-AUC denotes the area under the Precision Recall curve, as mentioned above, a suitable measure for performance in the presence of imbalanced data, independent of the decision boundary [13; 29].

$$Recall = \frac{TP}{TP+FN} \quad (3)$$

$$Precision = \frac{TP}{TP+FP} \quad (4)$$

We compare the following methods: random over-sampling (ROS), diversity-based over-sampling (D-ROS), SMOTE over-sampling (SMOTE), diversity-based SMOTE (D-SMOTE), under-sampling (RUS), diversity-based under-sampling (D-RUS). We evaluate the average performance measure (PR-AUC) of each over the four classifiers on each dataset. We also examine the statistical significance of the differences for every combination of classifiers and dataset using Wilcoxon signed-rank tests ($p < 0.05$) [30], which is a non-parametric test.

## 4.2 Experimental Design

We discuss the performance measure score (PR-AUC) by summarizing the results based for all classifiers and datasets.

### 4.2.1 Performance on classifiers.

The mean and standard error of PR-AUC for each combination of classifier and sampling methods are presented in Tables 2 to 4, using the original imbalance level, 5% imbalance level, and 1% imbalance level respectively, with the purpose of examining diversity-based performance as the imbalance level becomes more extreme.

For each imbalance level, we compare PR-AUC performance based diversity-based sampling methods against their respective non-diversity benchmark methods. In general, diversity-based sampling methods are statistically out-performing their benchmark in most cases: For D-SMOTE, 9 out of 12 diversity-based sampling methods outperform their benchmarks; for D-RUS, 11 out of 12 diversity-based sampling methods outperform their benchmarks. There are no cases where diversity-based methods perform statistically worse than their benchmark under any circumstances.

The superiority of diversity-based selection increases as the imbalance ratio becomes more extreme, i.e. when we decrease the original ratio to 1%. In other words, diversity-based sampling methods perform significantly better when the imbalance level is high in comparison with the benchmark re-sampling methods.

Similarly, as shown in Table 4, diversity-based under-sampling methods improve the PR-AUC performance measure score significantly for the majority of the classifiers. This can be understood as diversity-based selection providing a principle for under-sampling rather than just a randomness to keep the diversity and space distribution in the down-sampled majority group.

The diversity-based selection also helps SMOTE by reducing identical instances, which are generated using linear interpolations: identical instances do not contribute to the diversity of the dataset.

**Table 2:** PR-AUC results across classifiers with original imbalance levels. Bold numbers indicate the method performance is significantly better than the comparable methods (p < 0.05)

| Classifier | ROS | D-ROS | SMOTE | D-SMOTE | RUS | D-RUS |
|---|---|---|---|---|---|---|
| DT | 0.645±0.151 | 0.646±0.151 | 0.648±0.146 | **0.650±0.146** | 0.649±0.158 | **0.658±0.162** |
| GLM | 0.650±0.168 | 0.650±0.168 | 0.654±0.161 | **0.655±0.162** | 0.652±0.170 | **0.656±0.172** |
| KNN | 0.649±0.148 | 0.649±0.146 | 0.648±0.139 | **0.649±0.139** | 0.65±0.156 | **0.663±0.158** |
| RF | 0.621±0.145 | 0.621±0.145 | 0.626±0.139 | 0.626±0.139 | 0.619±0.151 | **0.627±0.15** |

**Table 3:** PR-AUC results across classifiers with 5% imbalance levels. Bold numbers indicate the method performance is significantly better than the comparable methods (p < 0.05).

| Classifier | ROS | D-ROS | SMOTE | D-SMOTE | RUS | D-RUS |
|---|---|---|---|---|---|---|
| DT | 0.675±0.136 | 0.674±0.137 | 0.664±0.141 | **0.667±0.143** | 0.691±0.151 | **0.709±0.159** |
| GLM | 0.662±0.155 | **0.665±0.153** | 0.661±0.158 | **0.662±0.159** | 0.671±0.166 | **0.679±0.168** |
| KNN | 0.671±0.135 | **0.673±0.135** | 0.657±0.138 | **0.660±0.138** | 0.678±0.153 | **0.703±0.158** |
| RF | 0.630±0.136 | 0.630±0.136 | 0.629±0.140 | 0.629±0.140 | 0.631±0.144 | **0.639±0.150** |

**Table 4:** PR-AUC results across classifiers with 1% imbalance levels. Bold numbers indicate the method performance is significantly better than the comparable classifiers (p < 0.05).

| Classifier | ROS | D-ROS | SMOTE | D-SMOTE | RUS | D-RUS |
|---|---|---|---|---|---|---|
| DT | 0.704±0.142 | **0.718±0.144** | 0.694±0.146 | **0.698±0.148** | 0.792±0.116 | 0.792±0.116 |
| GLM | 0.675±0.152 | **0.680±0.150** | 0.674±0.157 | **0.675±0.157** | 0.716±0.162 | **0.733±0.156** |
| KNN | 0.700±0.145 | **0.701±0.145** | 0.678±0.141 | **0.681±0.143** | 0.711±0.163 | **0.737±0.166** |
| RF | 0.641±0.141 | 0.641±0.141 | 0.639±0.142 | 0.639±0.142 | 0.647±0.154 | **0.672±0.162** |

**Table 5:** PR-AUC results across datasets with original imbalance levels. Bold numbers indicate the method performance is significantly better than the comparable methods (p < 0.05)

| Dataset | ROS | D-ROS | SMOTE | D-SMOTE | RUS | D-RUS |
|---|---|---|---|---|---|---|
| Wisconsin | 0.435±0.030 | 0.435±0.030 | 0.441±0.032 | 0.441±0.032 | 0.435±0.030 | **0.436±0.030** |
| Diabetes | 0.496±0.037 | **0.500±0.039** | 0.508±0.030 | **0.511±0.033** | 0.480±0.033 | **0.495±0.036** |
| KDD Control | 0.696±0.092 | 0.696±0.093 | 0.693±0.091 | 0.693±0.092 | 0.698±0.090 | 0.700±0.090 |
| Oil | 0.901±0.029 | 0.901±0.029 | 0.897±0.031 | 0.897±0.033 | 0.909±0.027 | **0.931±0.025** |
| Pima | 0.497±0.036 | **0.501±0.038** | 0.508±0.030 | **0.510±0.033** | 0.480±0.033 | **0.495±0.036** |
| Vehicle Buss | 0.529±0.027 | **0.530±0.027** | 0.542±0.028 | 0.542±0.028 | 0.528±0.029 | **0.534±0.032** |
| Vowel 0 | 0.782±0.035 | **0.783±0.035** | 0.781±0.035 | **0.783±0.035** | 0.782±0.031 | **0.785±0.033** |
| Vowel 2 | 0.779±0.030 | **0.780±0.032** | 0.778±0.030 | **0.781±0.031** | 0.806±0.035 | **0.809±0.036** |
| Wine Red Low vs High | **0.604±0.071** | 0.602±0.069 | 0.599±0.062 | 0.600±0.061 | 0.605±0.067 | **0.612±0.075** |
| Wine White Low vs High | 0.699±0.029 | 0.698±0.028 | 0.692±0.032 | **0.694±0.032** | 0.702±0.029 | **0.711±0.035** |

Table 6: PR-AUC results across datasets with 5% imbalance levels. Bold numbers indicate the method performance is significantly better than the comparable methods ($p < 0.05$)

| Dataset | ROS | D-ROS | SMOTE | D-SMOTE | RUS | D-RUS |
|---|---|---|---|---|---|---|
| Wisconsin | 0.444±0.034 | 0.445±0.034 | 0.449±0.035 | 0.449±0.035 | 0.440±0.033 | **0.447±0.041** |
| Diabetes | 0.540±0.048 | 0.542±0.041 | 0.531±0.042 | **0.534±0.042** | 0.521±0.046 | 0.525±0.047 |
| KDD Control | 0.696±0.086 | **0.700±0.088** | 0.697±0.087 | 0.697±0.087 | 0.712±0.086 | **0.745±0.088** |
| Oil | 0.901±0.029 | 0.901±0.029 | 0.897±0.031 | 0.897±0.033 | 0.909±0.027 | **0.931±0.025** |
| Pima | 0.540±0.048 | 0.542±0.041 | 0.522±0.046 | **0.523±0.046** | 0.521±0.046 | 0.525±0.047 |
| Vehicle Buss | 0.560±0.032 | **0.561±0.034** | 0.536±0.031 | **0.537±0.031** | 0.550±0.040 | **0.560±0.047** |
| Vowel 0 | 0.778±0.032 | 0.778±0.030 | 0.783±0.035 | 0.784±0.036 | 0.794±0.033 | 0.798±0.034 |
| Vowel 2 | 0.784±0.032 | **0.787±0.033** | 0.789±0.033 | 0.796±0.035 | 0.822±0.040 | 0.824±0.040 |
| Wine Red Low vs High | 0.638±0.075 | 0.635±0.077 | 0.622±0.069 | 0.623±0.068 | 0.691±0.100 | **0.697±0.100** |
| Wine White Low vs High | 0.713±0.037 | 0.712±0.037 | 0.703±0.031 | **0.704±0.032** | 0.715±0.034 | **0.769±0.058** |

Table 7: PR-AUC results across datasets with 1% imbalance levels. Bold numbers indicate the method performance is significantly better than the comparable methods ($p < 0.05$)

| Dataset | ROS | D-ROS | SMOTE | D-SMOTE | RUS | D-RUS |
|---|---|---|---|---|---|---|
| Wisconsin | 0.459±0.036 | 0.460±0.037 | 0.452±0.037 | **0.452±0.038** | 0.494±0.095 | 0.502±0.101 |
| Diabetes | 0.560±0.051 | **0.568±0.051** | 0.542±0.052 | 0.544±0.055 | 0.565±0.066 | **0.582±0.068** |
| KDD Control | 0.712±0.089 | **0.726±0.094** | 0.717±0.091 | 0.716±0.091 | 0.747±0.097 | **0.795±0.075** |
| Oil | 0.912±0.034 | 0.910±0.034 | 0.908±0.030 | 0.909±0.029 | 0.936±0.030 | **0.942±0.025** |
| Pima | 0.560±0.051 | **0.568±0.051** | 0.549±0.045 | 0.550±0.047 | 0.565±0.066 | **0.582±0.068** |
| Vehicle Buss | 0.580±0.044 | **0.583±0.048** | 0.570±0.038 | **0.571±0.038** | 0.653±0.085 | **0.667±0.077** |
| Vowel 0 | 0.820±0.040 | 0.823±0.043 | 0.799±0.035 | **0.802±0.035** | 0.862±0.057 | 0.864±0.056 |
| Vowel 2 | 0.821±0.038 | **0.829±0.039** | 0.820±0.037 | **0.828±0.041** | 0.869±0.041 | 0.872±0.046 |
| Wine Red Low vs High | 0.635±0.074 | 0.636±0.073 | 0.626±0.069 | 0.627±0.070 | 0.689±0.098 | **0.701±0.097** |
| Wine White Low vs High | 0.740±0.054 | **0.747±0.059** | 0.731±0.047 | **0.733±0.048** | 0.782±0.059 | **0.829±0.045** |

#### 4.2.2 Performance on datasets.

The mean and standard error of the PR-AUC score of all classifiers for each individual dataset and their corresponding sampling methods are presented in Tables 5 to 7, using the original imbalance level, 5% imbalance level, and 1% imbalance level respectively.

For each of these datasets, we compare diversity-based sampling methods with their non-diversity-based benchmarks. Diversity-based methods are statistically out-performing their benchmarks across most datasets: 54 out of 90 paired comparisons (60%) show significant improvement in PR-AUC. In particular, diversity-based under-sampling methods achieve higher means consistently across all datasets. There is a sole case where diversity-based selections underperform their benchmark (i.e. mean of D-ROS is statistically lower than ROS for Wine Red Low vs High).

We notice that diversity-based sampling methods performs better when the minority classes become sparser. For instance, the diversity-based selection does not outperform its benchmark for KDD Control data for the original imbalance ratio, but when the imbalance ratio is re-configured to 1% and 5% respectively, the performance of diversity-selection is significantly better for ROS and RUS. Similarly, for Wine White Low vs High data, the performance of diversity-selection statistically improves as the minority classes become sparser.

# 5 Case study

## 5.1 Experimental Design

The Austin dataset is split into 2/3 training and 1/3 testing to replicate the UK SORT experimental design. The PR-AUC performance measure score of the proposed model is compared to the PR-AUC performance measure of the original UK SORT model. The Austin dataset is re-sampled using ROS, D-ROS, SMOTE, D-DMOTE. We have also examined the hybrid re-sampling approaches by combining over-sampling and under-sampling with 25%, 50%, 75% ratios (OSUS, D-OSUS, SMOTEUS and D-SMOTEUS). For illustration, a 25% re-sampling ratio indicates that we under-sample the majority instances and over-sample the minority instances in a way the number of datapoints in the balanced training dataset is 25% of the original number of training datapoints. We employ the same logistic regression model (5) used in the UK SORT in order to provide and present comparable results.

$$\ln\left(\frac{M}{1-M}\right) = \beta_0 + \beta_1 x_1 + \beta_2 x_2 + \ldots + \beta_1 x_n \quad (5)$$

where M denotes the probability of a patient contracting morbidity within 30 days since the surgery, $\beta_0, \beta_1, \ldots, \beta_n$ denotes the coefficient parameters of our predictor variables, which is denoted by $x_1, x_2, \ldots, x_n$.

## 5.2 Experimental Design

The PR-AUC performance measure score of various re-sampling methods compared to UK SORT model is displayed in Table 8. The UK SORT model prediction is reproduced using coefficients published in [4]. For comparative purposes, the rest of our models only include features used in UK SORT [4] and are only built using logistic regression. We also build a logistic regression (benchmark) model using the Austin data without re-sampling methods. These two models work as baselines to show whether re-sampling methods can improve the predictability of post-surgery mortality. We notice that all of the diversified re-sampling methods return better results than the comparable non-diversity-based methods. Diversity based over-sampling (D-OS) produces the best result among all, with a 0.5% increase compared to the benchmark. The SMOTE re-sampling method does not perform particularly well. In this instance, the result can be explained by it relying on the generation of artificial data, thereby introducing more errors.

Table 9 shows an overview of our experimental results using various re-sampling methods and re-sampling ratios. We observe that in general diversity-based sampling methods perform better than their comparable non-diversity-based sampling methods. There is a sole case where D-SMOTEUS performs only at the same level as SMOTEUS. The best performing method is D-OSUS using a sampling ratio of 0.25. This can be explained as D-OSUS at a 0.25 re-sampling ratio requires an intensive amount of under-sampling and the least amount of over-sampling. During the diversified under-sampling process, we discard instances which have the least contribution to the diversity of our data. As there are a much larger pool of data for under-sampling than over-sampling, this naturally results in diversity under-sampling being a better indication of diversity in relation to diversity over-sampling. Overall, the best PR-AUC performance measure score of 93.7% is 1.4% higher than the benchmark PR-AUC score of 92.3%.

**Table 8:** PR-AUC results for logistic regression across different re-sampling methods compared to UK SORT model coefficient. Bold number indicates the best result in this table

| UK SORT Model | Benchmark | OS | D-OS | SMOTE | D-SMOTE |
|---|---|---|---|---|---|
| 0.922 | 0.923 | 0.925 | **0.928** | 0.921 | 0.922 |

**Table 9:** PR-AUC results for logistic regression across different re-sampling methods with 3 re-sampling ratios. Bold number indicates the best result in this table

| Re-sampling ratio | OSUS | D-OSUS | SMOTEUS | D-SMOTEUS |
|---|---|---|---|---|
| 0.75 | 0.924 | 0.928 | 0.919 | 0.920 |
| 0.5 | 0.922 | 0.924 | 0.922 | 0.923 |
| 0.25 | 0.921 | **0.937** | 0.922 | 0.922 |

**Table 10:** Model coefficient for the best performing logistic regression (D-OSUS 0.25 re-sampling ratios)

| Variable | Estimate | Std. Error | z value | Pr(>|z|) |
|---|---|---|---|---|
| (Intercept) | -3.892 | 0.503 | -7.734 | 1.04E-14 |
| ASAPS 2 | 1.902 | 0.324 | 5.868 | 4.41E-09 |
| ASAPS 3 | 1.932 | 0.320 | 6.024 | 1.71E-09 |
| ASAPS 4 | 3.637 | 0.441 | 8.231 | < 2e-16 |
| Emergency | 1.859 | 0.226 | 8.207 | 2.27E-16 |
| Severity = Intermediate | 1.526 | 0.467 | 3.266 | 0.001 |
| Severity = Major | 0.841 | 0.437 | 1.924 | 0.054 |
| Severity = Xmajor/Complex | 0.899 | 0.427 | 2.100 | 0.036 |
| Malignancy | 0.275 | 0.050 | 5.428 | 5.71E-08 |
| age_grp_1 | 0.597 | 0.244 | 2.443 | 0.015 |
| age_grp_2 | 0.764 | 0.272 | 2.803 | 0.005 |

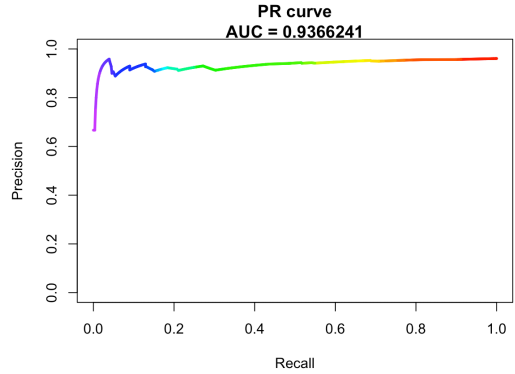

**Figure 1:** PR-AUC curve for the best performing logistic regression (D-OSUS 0.25 re-sampling ratio)

Figure 1 visualises the PR-AUC performance measure score curve. Table 10 shows the coefficients of the best performing logistic regression model (D-OSUS) with a sampling ratio of 0.25. All of the variables within this model are statistically significant at 5% . In comparison to the UK SORT model, our recorded intercept is -3.89, which is lower than the original UK SORT model. This is a good indication that on average, the patients recorded within our data experienced a lower mortality rate. Unsurprisingly, as patients' physical status progressively worsens (i.e. from healthy and normal (ASA-PS 1) to severe (ASA-PS 4)), the mortality risk increases. The mortality rate is higher for patients who have had emergency surgery, for those with comorbidity conditions (i.e. cancer) and of those with advanced age. We observe that surgery of intermediate severity results in a higher coefficient in comparison to complex surgery ("Xmajor/complex"), thereby increasing the risk of mortality. This is not in line with common intuition. However, this can possibly be attributed to the observation that the regression variables may not be perfectly independent. For example, there may be a correlation between the severity of surgery and the ASA-PS score, and therefore these incremental effects may have already been captured in other coefficients, such as in the "emergency" indicator.

# 6 Conclusions

In this study, our key objective has been to explore whether diversity-based selection improves the performance of re-sampling methods for constructing prediction models, specifically in the context of the UK SORT model. This is important, as an improvement in the SORT model or other future risk assessment tools has significant positive implications in assisting surgeons and patients to make informed decisions.

After validating the effectiveness of our diversity-based "drop-in" functionality with external datasets, we can conclude that it is effective in elevating the performance of the UK SORT by 1.4%. We also show that the extent of performance improvement is more prominent in situations where datasets have high and extreme imbalance ratios. This has significant potential implications, as it opens up an opportunity for future medical risk assessment tools to be built on rare medical events.

In this study, we have used the Solow-Polasky measure as a measure for diversity. Future research can investigate other diversity-based methods and incorporate them into the drop-in functionality, such as Mahalanobis Distance (MD). The integration of MD into the drop-in functionality may allow future studies to incorporate a probability density-based approach to bolster the extent of diversity.